\documentclass[letterpaper]{article} 
\usepackage{aaai24}  
\usepackage{times}  
\usepackage{helvet}  
\usepackage{courier}  
\usepackage[hyphens]{url}  
\usepackage{graphicx} 
\usepackage{natbib}  
\usepackage{caption} 
\usepackage{newfloat}
\usepackage{listings}
\DeclareCaptionStyle{ruled}{labelfont=normalfont,labelsep=colon,strut=off} 
\usepackage{amsmath}
\usepackage{cleveref}
\usepackage{amssymb}
\usepackage{enumitem}

\usepackage[ruled,vlined,linesnumbered]{algorithm2e}
\usepackage{url}      
\usepackage{booktabs} 
\usepackage{amsfonts} 
\usepackage{nicefrac} 
\usepackage{microtype}
\usepackage{xcolor}
\usepackage{nameref}
\usepackage{array}
\usepackage{epsfig}
\usepackage{mathtools}
\usepackage{epsfig}
\usepackage{multicol}
\usepackage{lipsum}

\nocopyright

\title{Task Planning for Object Rearrangement in Multi-room Environments}

\author {
    Karan Mirakhor\equalcontrib,
    Sourav Ghosh\equalcontrib,
    Dipanjan Das,
    Brojeshwar Bhowmick
}
\affiliations {
    Visual Computing and Embodied Intelligence Lab,\\ 
    TCS Research, Kolkata, India\\
    \{karan.mirakhor, g.sourav10, 
    dipanjan.da,
    b.bhowmick\}@tcs.com
}

\begin{document}

\maketitle
\vspace{-2mm}
\begin{abstract}
Object rearrangement in a multi-room setup should produce a reasonable plan that reduces the agent's overall travel and the number of steps. Recent state-of-the-art methods fail to produce such plans because they rely on explicit exploration for discovering unseen objects due to partial observability and a heuristic planner to sequence the actions for rearrangement. This paper proposes a novel hierarchical task planner to efficiently plan a sequence of actions to discover unseen objects and rearrange misplaced objects within an untidy house to achieve a desired tidy state. The proposed method introduces several novel techniques, including \textbf{(i)} a method for discovering unseen objects using commonsense knowledge from large language models, \textbf{(ii)} a collision resolution and buffer prediction method based on Cross-Entropy Method to handle blocked goal and swap cases, \textbf{(iii)} a directed spatial graph-based state space for scalability, and \textbf{(iv)} deep reinforcement learning (RL) for producing an efficient planner. The planner interleaves the discovery of unseen objects and rearrangement to minimize the number of steps taken and overall traversal of the agent. The paper also presents new metrics and a benchmark dataset called MoPOR to evaluate the effectiveness of the rearrangement planning in a multi-room setting. The experimental results demonstrate that the proposed method effectively addresses the multi-room rearrangement problem.
\end{abstract}

\vspace{-4mm}
\section{Introduction}

\label{sec:intro}
\par Organizing an untidy household according to user preferences is a challenging task that encompasses multiple aspects, including perception, planning, navigation, and manipulation \cite{batra2020rearrangement}. When an agent engages in multi-room object rearrangement, it must rely on sensor data and prior knowledge to devise a comprehensive plan that involves sequencing the object movement to achieve the desired tidy state. The specifications for this goal state can be defined using various modalities such as geometry, images, language, etc. \cite{batra2020rearrangement}.

\begin{figure}
    \centering \includegraphics[width=0.45\textwidth,height = 6cm]{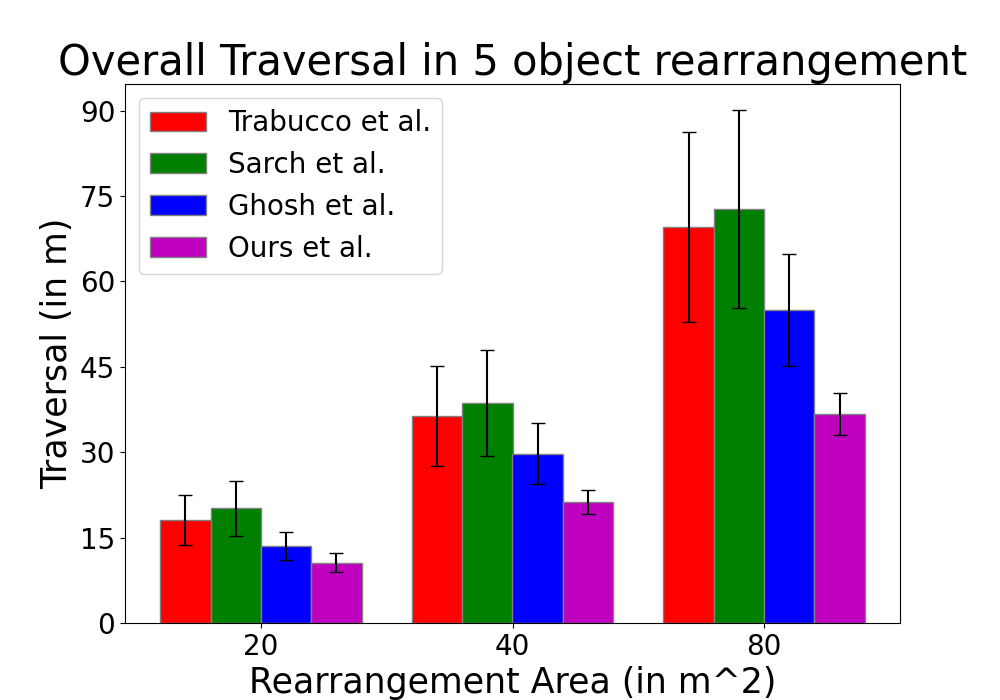}
    \caption{The graph shows the agent traversal for existing methods \cite{tidee, trabucco, Ghosh} v/s Ours with increasing rearrangement area highlighting need for efficient planning. The error bars show the standard deviation in the average traversal.}
    \label{fig:intro_graph}
\vspace{-7mm}
\end{figure}

The existing research on object rearrangement has predominantly concentrated on single-room setups, prioritizing perception and commonsense reasoning aspects. However, these studies often assume that navigation and manipulation abilities are already in place, and they do not emphasize the importance of efficient planning in the rearrangement process.
Methods such as \cite{kant2022housekeep,tidee} employ image or language-based commonsense reasoning to identify misplaced objects within the agent's egoview and then utilize sub-optimal heuristic planners for rearrangement. However, their reasoning based anomaly detectors fail to address cases where the goal position of an object is blocked or requires swapping with another misplaced object. Whereas methods such as \cite{vrr,csr,trabucco} focus on egocentric perception and employ image, graph, or semantic map-based scene representations to identify misplaced objects. They use greedy planners to sequence actions for rearrangement according to explicit user specifications. \cite{tidee} also performs user-specific room rearrangement by utilizing semantic relations to identify misplaced objects within the agent's egoview through exploration, followed by rearrangement using a heuristic planner. Most existing methods \cite{kant2022housekeep,tidee,trabucco,csr} explicitly explore the room to locate objects that are initially outside the agent's egoview, compensating for the partial information provided by the egocentric view. As shown in Fig~ \ref {fig:intro_graph}, recent methods \cite{tidee, trabucco, Ghosh} suffer from high traversal costs due to their sub-optimal planner, which becomes increasingly problematic as the size of the rearrangement space grows in multi-room settings.

Efficient planning is crucial for improving the effectiveness of rearrangement tasks by optimizing the sequence of actions and minimizing the time and effort required by the robot to achieve the desired goal state. In the context of rearrangement task planning, \cite{Ghosh} proposes a method that assumes complete room visibility from a bird's eye perspective. Their approach aims to overcome planning challenges, including the combinatorial expansion of rearrangement sequencing and swap case resolution, without requiring an explicit buffer.
However, their Euclidean distance-based reward does not minimize overall agent traversal during planning, as shown in Fig~\ref{fig:intro_graph}. Their state representation lacks scalability to large numbers of objects. Additionally, their parameter network suffers from two main issues: \textbf{(i)} predicting the goal location of non-blocked misplaced objects, which is already known from the goal state in rearrangement, leading to performance degradation. \textbf{(ii)} predicting the buffer location for swap cases without considering the object's geometry and the available free space, resulting in poor generalization. 

\begin{figure*}[t]
  \centering  \includegraphics[height = 5.5cm]{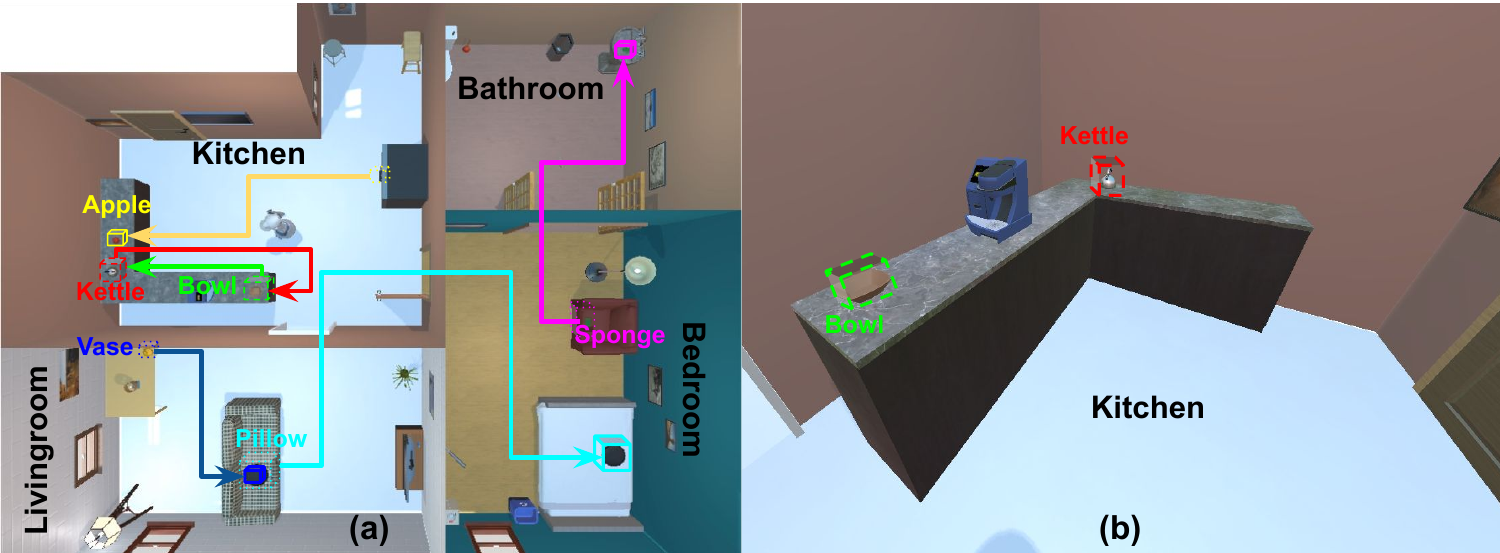}
  \vspace{-2mm}
   \caption{\textbf{(a)} shows the top down view of our Rearrangement task and \textbf{(b)} is the agent's initial egocentric view in the untidy current state for the same setup. The solid 3D bounding boxes indicate the desired goal state for all objects, while the dashed ones show the initial positions of visible objects in the untidy current state. The dotted 3D bounding boxes represent initial positions of unseen objects in the untidy current state. 
   The apple (yellow), an unseen object is inside the kitchen-fridge, while the vase (blue), pillow (pastel cyan) and sponge (magenta) is on the living-table, living-sofa and bedroom-chair respectively. There are two scenarios: a blocked goal case with the vase (blue) and pillow (pastel cyan) and a swap case between the bowl (green) and kettle (red).}
   \label{fig:teaser}
   \vspace{-5mm}
\end{figure*}

Furthermore, their reliance on ground truth object positions in both the current and goal states is impractical in real-life scenarios. In contrast, our focus is on a novel, practical and broader aspect of the object rearrangement problem in the multi-room setting, under partial observability using the egocentric camera view of the agent. Our main emphasis is on efficient task planning, which is necessary for effective rearrangement as illustrated in Fig~\ref{fig:intro_graph}. Task planning for efficient multi-room object rearrangement under partial observability as depicted in Fig~\ref{fig:teaser}, presents several significant challenges such as \textbf{(i)} uncertainty regarding the location of unseen objects due to partial observability, \textbf{(ii)} scalability to a large number of objects, \textbf{(iii)} combinatorial expansion of search space for the sequencing due to simultaneous object discovery (for unseen objects) and rearrangement, \textbf{(iv)} minimizing the overall traversal by the agent during simultaneous object discovery and rearrangement. \textbf{(v)} resolving blocked goal and swap cases (object collision) without explicit buffer.

In this paper, we present a novel hierarchical method for task planning  that aims to overcome the challenges mentioned earlier. Initially, our agent utilizes egocentric perception to explore the house once and capture the semantic and geometric configuration \cite{batra2020rearrangement} of objects and receptacles using any RGBD sensor \cite{brojo_hd} or SfM \cite{broj_sfm}, thus obtaining the goal state. The agent can also explore based on user commands \cite{broj_talk_vehicle}.  Following this, the objects within the rooms are shuffled to make an untidy current state. Our hierarchical method then divides the task planning problem into three components - the discovery of unseen objects, collision resolution, and planning - to minimize the agent's overall traversal while simultaneously conducting object search, collision resolution, and rearrangement.  \textbf{First}, we propose a novel commonsense knowledge-based Unseen Object Discovery Method using large language models (LLMs) \cite{RoBERTa,kant2022housekeep}, that leverages the object-room-receptacle semantics to predict the most probable room-receptacle for an unseen object. \textbf{Second}, we propose a novel Cross-Entropy Method (CEM) based collision resolution to produce buffer spaces for swap cases considering the objects' geometry and the size of receptacle-free spaces. \textbf{Third}, we use a Deep RL-based planner to produce an action sequence for simultaneous object search and rearrangement. To this extent, we define the Deep RL state space with a novel Directed-Graph-based representation that combines the geometric position of objects in the current state, the goal state along with the agent's initial position. The proposed representation effectively encodes the scene geometry, facilitating rearrangement planning and enabling scalability of the Deep RL state space to accommodate a large number of objects and scene invariant. By combining all the previously mentioned components in a thoughtful manner, we successfully address the combinatorial optimization problem in rearrangement. 

\noindent The major contributions of this paper are :
\begin{enumerate}[leftmargin=*,noitemsep,nolistsep]
\item To the best of our knowledge, this is the \textit{first end-to-end method} to address the task planning problem for multi-room rearrangement from egocentric view under partial observability, using a user-defined goal state.
\item A novel \textbf{Unseen Object Discovery Method} that leverages object-room-receptacle semantics using LLM to predict the most probable room-receptacle for an unseen object. 
\item Introducing \textbf{Cross-Entropy Method based Collision Resolution} to find buffer spaces for swap cases considering the objects' geometry and the size of free spaces. 
\item A new scalable and scene invariant \textbf{Directed State Graph} containing the geometric information about the agent and objects in the current and goal state as the Deep RL state.
\item Use of \textbf{Deep RL based planner} to overcome combinatorial expansion in rearrangement sequencing and, to optimize the overall traversal and the number of steps taken.
\item  we devise  a set of \textbf{Evaluation criteria} to gauge the efficacy of our method in terms of the number of steps taken and the overall agent traversal.
\item To address the inadequacies in existing benchmarks \cite{vrr} for evaluating multi-room task planning under partial observability, we introduce the \textbf{MoPOR - Benchmark Dataset}. 
\end{enumerate}
\section{Methodology}
\vspace{-2mm}
\subsection{Preliminaries}
\vspace{-1mm}
For our setup, we use the apartment scenes from ProcThor \cite{procthor}. To begin our multi-room rearrangement, we perform exploration \cite{tidee} once in the entire scene to capture the user-specified goal state. Using the RGB-D image and egomotion information at each step from Ai2Thor \cite{ai2thor}, the agent generates a 2D occupancy map ($\mathcal{M}^{\text{2D}}$) for navigation and a 3D map ($\mathcal{M}^{\text{3D}}$) to augment the positions of objects and receptacles in 3D to a global reference frame. A d-DETR \cite{d-detr} detector is used on each RGB image to obtain 2D bounding boxes and semantic labels for objects and receptacles. The corresponding 3D centroids are obtained using depth images, camera intrinsics and extrinsics. Using object segmentation \cite{pcl} on the point cloud of $\mathcal{M}^{\text{3D}}$ the agent records the corners of the 3D bounding boxes of each object in the goal state. At the end of the goal state exploration, we generate an object list $\textbf{O} = \{[\textbf{L}_i, \textbf{P}_i, \textbf{B}_i], i=1,2,..,N\}$ and a room-receptacle list $\textbf{R} = \{[\textbf{L}^R_i, \textbf{P}^R_i], i=1,2,..,N_R\}$. Here, $N$, $\textbf{L}$, $\textbf{P},\,\textbf{B}$ are the total numbers of objects, their semantic labels, 3D object centroids and the corners of their 3D bounding boxes respectively. $N_R$, $\textbf{L}^R$ and $\textbf{P}^R$ are the total numbers of receptacles, their semantic labels including the room name from Ai2Thor \cite{ai2thor},  and the 3D receptacle centroids respectively. After capturing the goal state, we randomly shuffle a few objects to make the room untidy and fork the agent at a random location in the room. In this untidy current state, the agent's knowledge is limited to the visible part of the scene in its egocentric view and thus only a set of objects $\textbf{O}^V = \{[\textbf{L}^V_i, \textbf{P}^V_i], i=1,2,..,N_V\}$ are captured using egocentric perception. $N_V$, $\textbf{L}^V$ and $\textbf{P}^V$ are the number of visible objects, their semantic labels and 3D object centroids respectively in the current state.
Additionally, the agent creates a 2D grid map of free receptacle spaces $\mathcal{M}^{\text{R}}$ in the current state using the instance segmentation mask from Ai2Thor along with the depth image and egomotion to aid in the collision resolution. Comparing $\textbf{O}$ with $\textbf{O}^V$ allows for determining only the semantics of unseen objects in the current state $\textbf{O}^{\overline{V}} = \{[\textbf{L}^{\overline{V}}_i], i=1,2,..,N_{\overline{V}}
\}$, where $N_{\overline{V}}$ is the number of unseen objects and $\textbf{L}^{\overline{V}}$ their semantic labels.
\begin{algorithm}
\caption{Algorithm for Task planner}
\label{alg:method}
\KwIn{Agent's egoview RGB-D \& egomotion}
\KwResult{Actions $\mathcal{A} = \{a_1,..,a_{N},done\}$}
Create $O, R, \mathcal{M}^\text{2D}, \mathcal{M}^\text{3D}$ from \text{Goal State}\; 
Create $O^{V}, O^{\overline{V}}, \mathcal{M}^\text{R}$ from \text{Current State}\;
\While{a is not done}{
$P^{\overline{V}R} \gets \textit{UODM (}O^{\overline{V}},R\text{)}$\;
\If{Collision between $O^{V}_i$ and $O^{V}_j$ }
{
\uIf{Swap case} 
{
$P^B_i \gets \textit{buffer for } O^{V}_i \textit{near} P^{V}_j$\; \label{swap_begin}
$P^B_j \gets \textit{buffer for } O^{V}_j \textit{near} P^{V}_i$\; \label{swap_end}
}
\uElseIf{$O^{V}_i$ blocks goal of $O^{V}_j$}
{
$P^B_j \gets P^{V}_j$\; \label{block_goal_begin}
}
\uElseIf{$O^{V}_j$ blocks goal of $O^{V}_i$}
{
$P^B_i \gets P^{V}_i$\;\label{block_goal_end}
}
}
$s \gets \textit{DSG (}O,P^B,O^V,O^{\overline{V}} \cup P^{\overline{V}R}\text{)}$ \;
$a = \underset{a\in \mathcal{A}}{\arg \max}\, Q_{\theta}(s,a)$\;
\uIf{$ a == {O^{V}_i}$}
{\textit{Pick-Place object} $O^{V}_i$\;}
\uElseIf{$ a == {O^{\overline{V}}_i}$}
{
\uIf{$O^{\overline{V}}_i$ \text{is found during traversal}}
{\text{go to} \ref{rem_rec_emp}\;} \label{inv_found_before}
\uElseIf{$O^{\overline{V}}_i$ \text{is found in} $P^{\overline{V}R}_i$}
{\textit{Pick-Place object} $O^{\overline{V}}_i$\;}\label{inv_found_at}
\uElse{\textit{Remove predicted receptacle from R}\;}\label{inv_not_found}
}
\textit{Remove receptacle/s without any} $O^{\overline{V}}$ \textit{from R}\;\label{rem_rec_emp}
\textit{Update} ($O^V, O^{\overline{V}}, R, \mathcal{M}^\text{R})$\;
\label{update}
}
\end{algorithm}
\vspace{-2.5mm}
\subsection{Overview}
\vspace{-1mm}   
Given \textbf{O}, \textbf{R}, $\textbf{O}^V$ and $\textbf{O}^{\overline{V}}$, the agent must efficiently plan a sequence of actions $\mathcal{A} = \{a_1, ...,\text{done}\}$ to discover $\textbf{O}^{\overline{V}}$ and simultaneously rearrange the misplaced objects in $\textbf{O}^V$ to their desired goal position in the current state. 
The Unseen Object Discovery Method leverages the object-room-receptacle relationship using $\textbf{O}^{\overline{V}}$ and \textbf{R} to predict the probable location $\textbf{P}^{\overline{V}R}$ for $\textbf{O}^{\overline{V}}$. The Collision Resolution and Buffer Management method, uses the locations and object bounding boxes from $\textbf{O}$ and $\textbf{O}^V$ along with $\mathcal{M}^{\text{R}}$ to detect object collision and predict the resolved goal location $\textbf{P}^B$ for the blocked goal and swap case objects. With \textbf{O}, $\textbf{O}^V$, $\textbf{O}^{\overline{V}}$, $\textbf{P}^{\overline{V}R}$, and $\textbf{P}^B$, the agent constructs a compact representation of the state space using the Directed State Graph. The Deep RL planner uses the state space to produce the most optimal action $a \in \mathcal{A}$ to either pick-place $\textbf{O}^{V}$ or search $\textbf{O}^{\overline{V}}$ with the objective of minimizing the overall traversal and the number of steps. 
\vspace{-2.5mm}
\subsection{Unseen Object Discovery Method}
\label{sec:UODM}
\vspace{-1mm}
In the context of multi-room rearrangement, the agent must identify unseen objects within the untidy current state, whether in the same room or different rooms, to plan actions effectively. For example, as shown in Fig 2, when the agent is in the kitchen, it cannot see items like the apple in the fridge-kitchen, or the vase (blue), pillow (pastel cyan), and sponge (magenta) in the living room and bedroom. To address this, we propose the Unseen Object Discovery Method (UODM), leveraging Large Language Models' (LLMs) commonsense knowledge to predict probable room-receptacles for unseen objects. However, LLMs may not always provide human-commonsense compliant predictions for untidy scenes, as detailed in Appendix\footnote{http://tinyurl.com/14043Appendix}.
Therefore, we leverage the semantic relationship between $\textbf{O}^{\overline{V}}$ and $\textbf{R}$ by finetuning their output embeddings from LLM using the AMT dataset \cite{kant2022housekeep}.
\vspace{-2.5mm}
\begin{equation}
\label{eq:SRTNloss}
\begin{aligned}
    L_{CE} = -\frac{1}{N_E}\sum^{N_E}\sum_{i=1}^{2} Y_{i} \log p_{i}\\
\end{aligned} 
\vspace{-1.5mm}
\end{equation}
\begin{equation}
\label{eq:SCNloss}
    L_{MSE} = \frac{1}{N_{SR}}\sum^{N_{SR}}_{i=1}(	\overline{\chi}_i - \chi_i)^2
\vspace{-0.75mm}
\end{equation}
Unlike previous methods such as \cite{tidee}, which focus solely on object-receptacle relationships, our method takes into account the object and room-receptacle semantic relationship to handle multi-room setups. We generate the RoBERTa embeddings $\textbf{E}^{\overline{V}R}$ for pairwise concatenated labels of unseen objects $\{\textbf{L}^{\overline{V}}_i\}_{i=1,2,..,N_{\overline{V}}}$ and room-receptacles $\{\textbf{L}^R_i\}_{i=1,2,..,N_R}$. As each object is paired with every receptacle, the total number of embeddings for all the object-room-receptacle (ORR) pairs is $N_E = N_{\overline{V}} \times N_R$. 
\begin{figure*}[t]
  \centering  \includegraphics[width=0.8\textwidth, height=7.5cm]{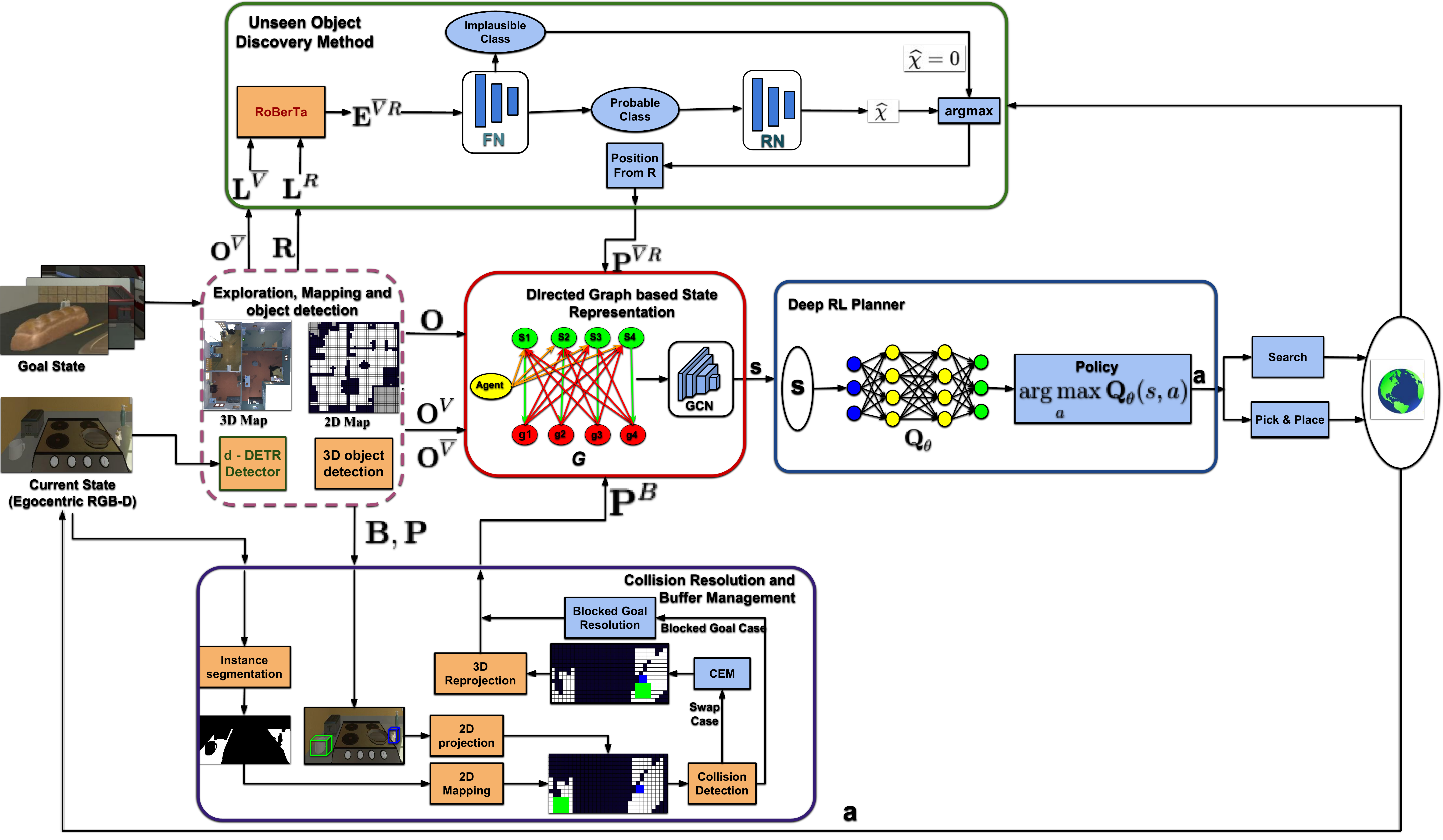}
  \vspace{-3mm}
   \caption{Overall hierarchical pipeline of our proposed method.}
   \label{fig:Pipeline-1}
   \vspace{-6mm}
\end{figure*}
We devise a two-step approach to enhance the accuracy and reduce the search space for finding the exact ORR from the $N_E$ pairs. \textbf{First}, we use an MLP-based Filter Network (FN) to predict the probability $\{p_i\}_{i=1,2}$ for ORR from $N_E$. We train this network using a Cross-Entropy Loss ($L_{CE})$ on the ground truth class labels $\{Y_i\}_{i=1,2}$ for each ORR in the dataset, as shown in Eq~\ref{eq:SRTNloss}. Here, the class labels $\{i=1:\textit{Probable Class},\,2:\textit{Implausible Class}\}$ indicate the probability of finding a misplaced object at a given room-receptacle. In the \textbf{second} step, we use a regression-based Ranking Network (RN) to estimate the probability scores $\{ \overline{\chi}_i\}_{i=1,2,...N_{SR}}$ for embeddings of probable class, with $N_{SR}$ representing the total number of such embeddings. We use a fully connected MLP, and train this network using MSE Loss ($L_{MSE}$) as shown in Eq~\ref{eq:SCNloss}, with respect to the ground truth probability scores $\{\chi_i\}_{i=1,..N_E}$ from human annotations. The room-receptacles with the highest scores from RN are selected as the probable room-receptacles for the unseen objects $O^{\overline{V}}$. 
Finally, the predicted room-receptacle's position $\{\textbf{P}^{\overline{V}R}_i\}_{i=1,..N_{\overline{V}}}$ from the goal state is used as the location for $\textbf{O}^{\overline{V}}$ in the current state, as receptacles are static in the scene. 

To prevent fruitless searches, we implement simple strategies as shown in Line~\ref{inv_found_before}:\ref{update}. Line~\ref{rem_rec_emp} exemplifies a situation where the agent comes across a receptacle on its path that does not contain any $O^{\overline{V}}$. As a result, the agent excludes it from future search attempts.
\vspace{-3mm}
\subsection{Collision Resolution and Buffer Management}
\vspace{-1.5mm}
To ensure the robustness of our task planning in collision scenarios, we need to identify and resolve the blocked goal and swap cases by leveraging the geometry of objects ($B$) and the size of free spaces in receptacle occupancy map ($\mathcal{M}^{\text{R}}$).
We identify the collision cases between objects in $\textbf{O}^V$, using $B_i, B_j \in B$ for each pair of objects and $\mathcal{M}^{\text{R}}$. We project $B_i, B_j$ onto $\mathcal{M}^{\text{R}}$ and find the maximum size rectangle $B^M_i$ and $B^M_j$ that encloses their projections respectively. Additionally, we require the projected goal positions $p^M_i, p^M_j$ of the two objects in $\mathcal{M}^{\text{R}}$. To identify the type of collision case, we check \textbf{(i)} $B^M_i$ $\cap$ \{$B^M_j \rightarrow p^M_j\} (B^M_j$ centred at $p^M_j) \neq \phi$ and \textbf{(ii)} $\{B^M_i \rightarrow p^{M}_i\}$ $\cap$ $B^M_j \neq \phi$.Three cases arise based on these conditions: \textbf{(i)} If none of the conditions are true, there is no collision, \textbf{(ii)} If only one of the conditions is true, it is a blocked goal case, and \textbf{(iii)} If both conditions are true, it is a swap case. 

To resolve swap cases, we find a buffer space in a time bound and effective manner, using a stochastic optimization over $p_k \in \mathcal{M}^{\text{R}}$ that maximizes the objective function $\mathcal{F}(p_k)$ in Eq~\ref{CEM}\footnote{shows an objective function to find a buffer for $B^M_i$ near $\{B^M_j \rightarrow p^M_j\}$}. To perform this optimization, we use the Cross-Entropy Method (CEM) - a simple derivative-free optimization algorithm, which is parallelizable and moderately robust to local optima \cite{cem}. CEM samples a batch of $N_{f}$ points from the free spaces in $\mathcal{M}^{\text{R}}$ at each iteration. Then, it fits a Gaussian distribution to the best $M_f$ $\leq$ $N_{f}$ samples based on $\mathcal{F}(p_k)$. Further samples for the next batch of $N_f$ are taken from this Gaussian.
\begin{equation}
\label{CEM}
\mathcal{F}(p_k) = 
\begin{dcases}
    e^{-\mathcal{E}(p_k,p^M_j)},& \text{if } \{B^M_i \rightarrow p_k\} \cap \{B^M_j \rightarrow p^M_j\} = \phi\\
    0,              & \text{otherwise}
\end{dcases}
\end{equation}
Here, $\mathcal{E}(p_k,p^M_j) = \|(p_k-p^M_j)\|_2$. 
The buffer locations obtained from CEM are used to update $P^B$ for swap case objects as shown in Line~\ref{swap_begin}:\ref{swap_end}. Once a swap object is placed in its buffer space, it is considered a blocked goal case. To resolve a blocked goal case, the goal blocked object is made static temporarily, unless the goal blocking object vacates its goal position. This is done by updating $P_B$, as in Line~\ref{block_goal_begin}:\ref{block_goal_end}. 
\vspace{-6mm}
\subsection{Rearrangement planner}
\vspace{-1mm}
Rearrangement planning is a long-horizon problem that suffers from the combinatorial expansion with increasing number of objects.
To this end, we employ data-driven Q-learning \cite{qlearning} that leverages the Bellmann principle \cite{bellmann} for achieving optimality in long horizon problems. Moreover, to aid the planner we use the Directed spatial graph as the state space which allows for scene invariance and scalability to a large number of objects.
\vspace{-2mm}
\subsubsection{Directed State Graph}
We present a directed spatial graph ($\mathcal{G}=\{V,E\}$) to create a concise state space representation to aid the planner in efficient rearrangement sequencing. The edges in this directed graph represent all the feasible paths for rearrangement completion, disregarding the redundant information present in an undirected graph, thereby improving the training efficiency (refer Appendix$^{2}$).  The nodes $V$ (shown in Fig~\ref{fig:Pipeline-1}) contain \textbf{(i)} the agent node with 3D position of the agent, \textbf{(ii)} the source nodes with current object positions and labels from $\{O^V,\{O^{\overline{V}} \cup P^{\overline{V}R}\}\}$ and \textbf{(iii)} the goal nodes with the goal object positions and labels from $O$. Unlike \cite{Ghosh}, we include the agent node in the Deep RL state space, as it enables the planner to effectively select the initial action based on the agent's position with respect to the objects'. The directed edges of the graph connect: \textbf{(i)} the agent node to every source node, \textbf{(ii)} the source node of each object to its respective goal node, and \textbf{(iii)} the goal node of each object to the current node of every other object. The edge attributes $E_c = \{\mathcal{D}(p^G_i, p^G_j)_{i \neq j}\}$ include the length of the shortest collision free path $\mathcal{D}(p^G_i, p^G_j)_{i \neq j}$ from the node position $p^G_i$ to connected node position $p^G_j$. $\mathcal{D}(p^G_i, p^G_j)_{i \neq j}$ is computed using BFS algorithm between the 2D projections of $p^G_i, p^G_j$ on $\mathcal{M}^{\text{2D}}$. For unseen objects in the current state, the source object nodes in $G$ are augmented with $P^{{\overline{V}}R}$ from UODM. Similarly, for blocked goal and swap cases, the goal nodes of such objects are updated with $P^{B}$ from Collision Resolution and Buffer Method. This graph representation helps the Deep RL state space to understand the geometric information of the rearrangement context. We use a \textit{Graph Convolution Network (GCN)} to generate meaningful graph embedding from $\mathcal{G}$, that enables the Deep RL state space to remain scalable and scene invariant. 
\vspace{-3mm}
\subsubsection{Deep RL based Planner}
The task planner should efficiently plan a sequence of actions to simultaneously \textbf{(i)} rearrange visible objects and \textbf{(ii)} search for unseen objects at probable receptacles. This reduces the agent's overall travel time as \textbf{(i)} it eliminates the requirement of explicit exploration to find unseen objects and \textbf{(ii)} the rearrangement of visible objects inevitably leads to the discovery of some unseen objects.
To accomplish these objectives, we utilize a Conservative Q-Learning based on reinforcement learning, which is similar to the approach presented in \cite{cql}. The state space for Deep RL is defined as $s = Z_p$ and action $a \in \mathcal{A}$ denotes the selected object in $\textbf{O}^{V}$ or $\textbf{O}^{\overline{V}}$. Our approach follows the principles of a \textit{Markov Decision Process (MDP)}, where a reward $r(s,a)$ is received at each time step $t$ for selecting $a$ from the policy $\pi(a|s) = \underset{a\in \mathcal{A}}{\arg \max}\, Q_{\theta}(s,a)$, that moves the agent from the current state $s$ to the next state $\bar{s}$. So according to Bellman equation Eq~\ref{TD}, our objective is to minimize the temporal difference error to get the desired action.
\begin{equation}
\label{TD} 
\begin{aligned}
L_{TD} = 
\underset{(s,a,\bar{s})\leftarrow R_B}{\frac{1}{2} \, \mathbb{E}}[(r(s,a) + \gamma\,\underset{\bar{a}\in \mathcal{A}}{\max}\,Q_{\bar{\theta}}(\bar{s},\bar{a}) - Q_{\theta}(s,a))^{2}]    
\end{aligned}
\end{equation}
Here, ${\theta}$ and $\bar{\theta}$ are the parameters of the Q-Network and target Q-Network respectively and $R_B$ is the replay buffer. The Q-Network tends to overestimate the policy value, which is undesirable and affects the sampling efficiency (refer Appendix$^{2}$). To prevent this, we employ the  Conservative Q-Learning technique similar to that used in \cite{cql}. This ensures that the expected value of a policy under the learned Q-function provides a lower-bound estimate of its true value. Therefore, the combined loss obtained on adding the conservative lower bound loss to Eq~\ref{TD} is shown in Eq~\ref{CQL} 
\begin{equation}
\label{CQL} 
\begin{aligned}
L_{CQL} = \underset{\underset{a \sim \pi(a|s)}{s \leftarrow R_B}}{\alpha\,(\mathbb{E}}\,[Q_{\theta}(s,a)] - \underset{s,a \leftarrow R_B}{\mathbb{E}}\,[Q_{\theta}(s,a)]) + L_{TD}  
\end{aligned}
\end{equation}
Here $\alpha \geq 0 $ is tradeoff factor between the conservative loss term and the TD error. When it comes to Long Horizon planning, using the sparse reward is not an efficient method for training Deep RL, as noted in \cite{clement2021reinforcement}. 

Therefore, we compute a hierarchical dense reward structure which has three components. (i) \textit{Rearrange-able Object Reward :} If the agent selects a misplaced object (an object whose position in $\textbf{O}$ and $\textbf{O}^{V}$ differs by at least 10cm), it receives a reward equal to the negative path length required for the agent to complete the pick-and-place action for that object. (ii) \textit{Static Object Reward :} Penalizes the agent for moving non-rearrangeable or static objects, thus preventing redundant moves. This reward also prevents the Deep RL planner from choosing the goal-blocked object which is temporarily set as static by making its goal position same as its current position, unless the goal-blocking object vacates its goal position. (iii) \textit{Completion Reward :} If the agent correctly rearranges only the misplaced objects, it gets a high positive reward. 
(iv) \textit{Collision Resolution :} It prioritizes the goal-occupied objects (objects which occupy the goal location of other objects) to move first instead of the goal-blocked objects. 
More details regarding the reward structure and the choice of rewards, can be found in the supplementary material.

We employ an off-policy technique to train our Conservative Q-Learning method using a diverse set of rearrangement configurations, which is similar to the approach proposed by \cite{kalashnikov2018qt}. To balance exploration and exploitation, we use the $\epsilon$ greedy method, as described in \cite{kalashnikov2018qt}. To ensure stable Off-policy training, we update $\bar{\theta}$ weights using polyak averaging on $\theta$, which is similar to \cite{bester2019multi}.
\vspace{-2mm}
\section{Experiment}
\label{sec:exp}
In this section, we describe the dataset, metrics, and detailed results of our proposed method and its modules, in addressing the multi-room rearrangement problem.
\vspace{-3mm}
\subsection{Unseen Object Discovery dataset}
\vspace{-1mm}
The AMT dataset \cite{kant2022housekeep} consists of 268 object categories present in 12 distinct rooms and 32 receptacle types. Each object-room-receptacle (ORR) pair is evaluated by 10 annotators who rank them into one of three classes: correct (positively ranked), misplaced (negatively ranked), or implausible (not ranked). By calculating the mean inverse of the ranks assigned to each ORR, we obtain the ground-truth scores. For our specific problem, our preference order for ORRs is as follows: misplaced class, correct class, and lastly, the implausible class.
Hence, we re-label the classes and their scores as (i) misplaced and correct class $\rightarrow$ probable class, while (ii) implausible class remains the same. 
\vspace{-3mm}
\subsection{Benchmark Dataset for Testing - MoPOR}
\vspace{-1mm}
The existing benchmark dataset, RoomR \cite{vrr}, is utilized to evaluate rearrangement policies across different room scenarios. However, it has certain limitations. It restricts the number of objects to a maximum of 5 and does not allow object placement within another receptacle, nor does it include blocked goal or swap cases. Consequently, it cannot adequately assess planning aspects such as the number of steps, agent traversal, blocked goals, or swap cases.
To overcome these limitations, we introduce MoPOR, a new benchmark dataset designed specifically for testing task planners in Ai2Thor. MoPOR encompasses a diverse collection of single-room scenes from iThor and multi-room scenes from ProcThor \cite{procthor} \footnote{Every scene has 4 types of rooms - living room, bedroom, bathroom, kitchen}. It supports up to 108 object and receptacle categories. This dataset enables a wide range of rearrangement scenarios involving up to 40 objects. Furthermore, MoPOR includes random partial observability cases, object placement within receptacles in the current state, as well as blocked goal and swap cases.
Moreover, it's worth noting that object placement configurations in MoPOR impact the effectiveness of sub-optimal planning policies in terms of agent traversal. 
Further details on the distribution of objects, rooms and receptacles in Appendix.

\vspace{-2mm}
\subsection{Training}
\vspace{-1mm}
\label{sub:Training}
The training details of UODM and DSG with Deep RL planner are available in the Appendix$^{2}$.
\vspace{-2mm}
\subsection{Evaluation Criteria}
\vspace{-1mm}
Metrics in \cite{vrr} do not highlight the efficacy of a task planner to judge efficient sequencing to reduce the number of steps taken or the agent traversal during rearrangement. For a fair evaluation of our method, and comparison against the existing methods and ablations, we define the following criteria:
\begin{itemize}[leftmargin=*,noitemsep,nolistsep]
    \item \textbf{SRN} : \textbf{S}uccess \textbf{R}atio measures rearrangement episode success and efficiency by combining the binary success rate (S) and the \textbf{N}umber of steps ($N_S$) taken by the agent to rearrange a set number of objects ($N$). A higher SRN indicates a more efficient and successful rearrangement episode, as it implies a lower $N_S$ for a given $N$.
    (\textbf{SRN} = $S \times N \slash N_{S}$)
    \item \textbf{EOD}: \textbf{E}fficiency in unseen \textbf{O}bject \textbf{D}iscovery is the ratio of the number of unseen objects initially ($N_{\overline{V}}$) with respect to the number of attempts to search ($N_{S\overline{V}}$). A higher \textbf{EOD} shows a lower $N_{S\overline{V}}$ for a given $N_{\overline{V}}$ indicating a more efficient search to find unseen objects.(\textbf{EOD} = $N_{\overline{V}} \slash N_D$)
    \item \textbf{TTL}: \textbf{T}otal \textbf{T}raversal \textbf{L}ength metric shows the total distance traversed by the agent during the successful completion of a rearrangement episode. In an identical test configuration, a lower \textbf{TTL} indicates a more efficient rearrangement sequencing .
\end{itemize}

\begin{table*}[t]
  \begin{center}
  \centering
  \resizebox{1.0\textwidth}{0.1\textwidth}{
  \begin{tabular}
  {>{\centering\arraybackslash}c>{\centering\arraybackslash}c>{\centering\arraybackslash}c>{\centering\arraybackslash}c>{\centering\arraybackslash}c>{\centering\arraybackslash}c>{\centering\arraybackslash}c>{\centering\arraybackslash}c>{\centering\arraybackslash}c>{\centering\arraybackslash}c>{\centering\arraybackslash}c>{\centering\arraybackslash}c>{\centering\arraybackslash}c>{\centering\arraybackslash}c>{\centering\arraybackslash}c>{\centering\arraybackslash}c>{\centering\arraybackslash}c>{\centering\arraybackslash}c>{\centering\arraybackslash}c>{\centering\arraybackslash}c 
  }
    \toprule
\begin{tabular}[c]{@{}c@{}}Number\\of Objects\end{tabular} &\begin{tabular}[c]{@{}c@{}}Visible\\Objects\end{tabular} & \multicolumn{2}{c}{\begin{tabular}[c]{@{}c@{}}Unseen\\Objects\end{tabular}} & \begin{tabular}[c]{@{}c@{}}Swap\\Case\end{tabular}& \multicolumn{3}{c}{\textbf{Ours-GT}} & \multicolumn{3}{c}{\textbf{Ours}} &\multicolumn{3}{c}{Ours-RS}
  &  \multicolumn{3}{c}{Ours-GE}&  \multicolumn{3}{c}{Ours-HP}\\
    \midrule
    &  &\textbf{P.O} & \textbf{F.O}&&\textbf{SRN}  & \textbf{EOD} & \textbf{TTL} & \textbf{SRN}  & \textbf{EOD} & \textbf{TTL} &\textbf{SRN}  & \textbf{EOD} & \textbf{TTL} &\textbf{SRN}  & \textbf{EOD} & \textbf{TTL} &\textbf{SRN}  & \textbf{EOD} & \textbf{TTL} \\
    \midrule    
    $10$ &6 &4 &0 &2 &\textbf{0.59} & \textbf{0.55} &\textbf{37.03} &\textbf{0.41} & \textbf{0.53} & \textbf{37.82} &0.19 &0.20 &60.72 &0.15 &0.13 &70.09  &\textbf{0.41} & \textbf{0.53} & {45.82}\\
    
    &6 &0 &4 &2 &\textbf{0.55} &\textbf{0.51} &\textbf{38.25}&\textbf{0.39} &\textbf{0.48} &\textbf{40.27} &0.13 &0.15 &65.47 &0 &NC &NC &\textbf{0.39} &\textbf{0.48} &{50.27}\\
    
    \midrule
    $20$&12 &8 &0 &4 &\textbf{0.61} & \textbf{0.60} & \textbf{66.94}& \textbf{0.43}& \textbf{0.56} & \textbf{69.29} &0.22 &0.24 &89.45 &0.19 &0.17 &101.68  &\textbf{0.43}& \textbf{0.56} & 82.42\\
    
    &12 &0 &8 &4&\textbf{0.58} &\textbf{0.55} &\textbf{68.95} &\textbf{0.40} &\textbf{0.53} &\textbf{72.38} &0.17 &0.18 &98.37 &0 &NC &NC  &\textbf{0.40} &\textbf{0.53} &89.67\\
    
    \midrule
    $30$ &18 &12 &0 &6 &\textbf{0.64} & \textbf{0.67} & \textbf{90.25}& \textbf{0.45}& \textbf{0.62} & \textbf{94.76} &0.27 &0.28 &120.45 &0.24 &0.19 &131.92  &\textbf{0.45}& \textbf{0.62} & 109.42\\
    
    &18 &0 &12 &6&\textbf{0.61} &\textbf{0.61} &\textbf{95.71} &\textbf{0.43} &\textbf{0.58} &\textbf{98.58} &0.23 &0.22 &132.16 &0 &NC &NC  &\textbf{0.43} &\textbf{0.58} &118.72\\

    \bottomrule
  \end{tabular}}
  \vspace{0.2mm}
  \caption {\small{Results for Multi-room Rearrangement with comparison against Baselines. Here, \textbf{P.O.} indicates partially occluded cases i.e. objects which are outside the field of view presently and \textbf{F.O.} stands for fully occluded cases i.e. objects placed inside closed receptacles. \textit{Ours-GE} does not handle \textbf{F.O.} cases, therefore its \textbf{EOD} is non-computable (\textbf{NC}) due to division by zero. The table shows that \textit{Ours-GT} and \textit{Ours} outperform the other baselines in terms of the evaluation criteria.}}
  \label{tab:multi_Room}
  \end{center}
  \vspace{-6mm}
\end{table*}
\vspace{-2mm}
\subsection{Baselines}
\vspace{-1mm}
\label{subsec:Baselines}
We ablate our method against ground-truth perception, various methods for object search and different planners. To study the effect of erroneous perception on our method, we use \textbf{(i)} \textbf{\textit{Ours-GT}} with ground-truth perception - 3D object detection and instance segmentation mask from Ai2Thor \cite{ai2thor}. To understand the importance of UODM in our method, we replace it by a  random search policy in \textbf{(ii)} \textbf{\textit{Ours-RS}}, which predicts probable receptacles for unseen objects with uniform probability and a greedy exploration strategy \cite{ANS} in \textbf{(iii)} \textbf{\textit{Ours-GE}} that optimizes for map coverage to discover all the unseen objects. To gauge the efficacy of our planner we replace the Deep RL planner in our method with a heuristic planner in \textbf{(iv)} \textbf{\textit{Ours-HP}} that greedily selects an action with the shortest agent traversal to complete the object pick-place.    
\vspace{-2.5mm}
\subsection{Results}
\label{sec:quantr}
\subsubsection{Ablations}
\label{multi_room}
The state-of-the-art methods dealing with user-defined goal state do not demonstrate their results in multi-room rearrangement task. Due to this limitation, we gauge the performance of our method in Tab~\ref{tab:multi_Room}, by comparing it against the set of baselines in a multi-room setting on MoPOR - Benchmark Datset. Tab~\ref{tab:multi_Room} indicates that our method is scalable to a large number of objects, with consistently increasing \textbf{SRN} values for swap cases and both scenarios of partial observability (\textbf{P.O.} : objects outside the field of view, \textbf{F.O.} : objects inside closed receptacles). In addition, the consistently high \textbf{SRN} for a growing number of swap cases indicates that \textit{Ours} and \textit{Ours-GT} can effectively handle swap cases, using the Cross-entropy based method for buffer prediction. The gradual increase in \textbf{SNR} and \textbf{EOD} with the increase in number of objects can be attributed to the fact that rearrangement of visible objects and the search for some unseen objects, indirectly aids in finding other unseen objects.
As anticipated, \textit{Ours-GT} results in significantly better \textbf{SNR}, \textbf{EOD} and \textbf{TTL}, compared to Ours and all the baselines, because it uses ground-truth object detection and labelling.
\vspace{-0.4mm}
\par The variation in the results between Ours and the baselines primarily arises from their approaches to handle partial observability cases, since Ours and the baselines employ the same buffer prediction method for swap cases. In both the cases of partial observability ($\textbf{P.O.}$ $\&$ $\textbf{F.O.}$), Ours performs significantly better than Ours-GE, Ours-RS and Ours-HP in terms \textbf{SRN}, \textbf{EOD} and \textbf{TTL}. This is due to the efficacy of the Unseen Object Discovery Method (UODM) and the efficient planning of Deep RL during simultaneous object search and rearrangement. Ours-GE incurs a high traversal cost in terms of \textbf{TTL} because it explicitly explores the entire apartment to find the \textbf{PO} objects. Moreover, Ours-GE fails to address the \textbf{FO} cases (\textbf{SRN = 0}) because the greedy exploration policy \cite{ANS} in terms of map coverage does not include opening and closing receptacles to find \textbf{FO}. Whereas, Ours-RS randomly visits receptacles to discover \textbf{PO} or \textbf{FO} cases, which again increases \textbf{TTL}. In contrast, our approach performs similarly in both cases of partial observability because UODM comprehends a semantic relationship between an object and any type of room-receptacle - rigid or articulated. To gauge the performance of the exploration strategy for object search in terms of \textbf{EOD}, we consider each newly generated location or a set of navigational steps from the exploration policy as a search attempt. More number of attempts to search in Ours-GE and Ours-RS lead to a lower \textbf{EOD} as well as \textbf{SRN}. We observe that Ours-RS performs slightly better than Ours-GE in terms of \textbf{EOD} and \textbf{TTL} for \textbf{PO}, because Ours-RS interleaves object search and rearrangement, rather than doing an explicit exploration strategy for finding objects. Ours-HP\footnote{Ours-HP has similar UODM and Collision Resolution and Buffer Management Method} only shows a higher \textbf{TTL} compared to Ours for both \textbf{PO} and \textbf{FO} cases in partial observability due to the greedy planning method.

Please refer to the appendix$^{2}$ to find results for the analysis of the choice of hyper-parameters for each of our learning-based modules. 

\vspace{-2mm}
\subsubsection{Comparison against state-of-the-art methods for Room-Rearrangement}
The existing methods train and show results on a single-room rearrangement task. Therefore, for a fair comparison, we compare our method against them in a single-room setting on MoPOR - Benchmark Dataset and RoomR \cite{vrr}. Please refer to the appendix$^{2}$ to find the detailed comparison tables and discussion.

\vspace{-2mm}
\subsection{Qualitative Results}
\vspace{-1mm}
To show the qualitative results of our method in multi-room rearrangement, we have created multiple test scenario videos to highlight the robustness of our method. We also evaluate our method in a new environment- Habitat, as highlighted in our supplementary video\footnote{http://tinyurl.com/SuppVideo}.
This transfer does not require any additional training for our UODM or Deep RL planner. This demonstrates how our method excels in seamless sim-to-sim transfer, reinforcing its suitability for deployment in real-world scenarios.
Please refer the supplementary video$^{6}$.

\vspace{-2mm}
\section{Limitations}
\vspace{-1mm}
Our approach is not capable of identifying unseen objects that are occluded due to clutter on receptacles (for e.g. a spoon may become occluded, if bread, box, lettuce etc. is placed before it). Our method also assumes the availability of perfect motion planning and manipulation capabilities.
\vspace{-2mm}
\section{Conclusion}
\vspace{-1mm}
This paper introduces a novel task planner designed for tidying up an apartment while dealing with partial observability, which can be adapted to various situations and generates a sequence of actions that minimizes the agent's overall traversal and the number of steps taken during simultaneous unseen object discovery and rearrangement. In future work, we plan to explore the deployment of our method in real-world.

\bibliography{aaai24}

\end{document}